\documentclass[11pt,letterpaper]{article}
\usepackage{emnlp2017}
\usepackage{times}
\usepackage{latexsym}
\usepackage{graphicx}
\usepackage{caption}
\usepackage{subcaption}
\usepackage{booktabs}
\usepackage{pbox}
\usepackage{multirow}
\usepackage[flushleft]{threeparttable}
\usepackage{amsmath}

\emnlpfinalcopy

\def\emnlppaperid{***}

\title{Social Media-based Substance Use Prediction}

\author{Tao Ding$^*$, Warren K. Bickel$^\dagger$, \and Shimei Pan$^*$ \\
       $*$ Department of Information Systems \\
       University of Maryland, Baltimore County \\ 
       \texttt{\{taoding01,shimei\}@umbc.edu}\\
       $\dagger$Addiction Recovery Research Center\\
       Virginia	Tech Carilion Research Institute\\
       \texttt{wkbickel@vtc.vt.edu}
       }

\begin{document}

\maketitle
\begin{abstract}
In this paper, we demonstrate how the state-of-the-art machine learning and text mining techniques can be used to build effective social media-based substance use detection systems.  Since a substance use ground truth is difficult to obtain on a large scale, to maximize system performance, we explore different feature learning methods to take advantage of a large amount of unsupervised social media data. We also demonstrate the benefit of using multi-view unsupervised feature learning to combine heterogeneous user information such as Facebook ``likes" and  ``status updates"  to enhance system performance.  Based on our evaluation, our best models achieved 86\% AUC for predicting tobacco use,  81\% for alcohol use and 84\% for drug use, all of which significantly outperformed existing methods. Our investigation has also uncovered interesting relations between a user's social media behavior (e.g., word usage) and substance use.
\end{abstract}

\section{Introduction}
A substance use disorder (SUD) is defined as a condition in which recurrent use of substances such as  alcohol, drugs and tobacco causes clinically and functionally significant impairment in an individual's daily life ~\cite{samhsa2015}. According to the 2014 National Survey on Drug Use and Health, 1 in 10 Americans age 12 and older had a substance use disorder. Substance use also costs Americans more than \$700 billion a year in increased health care costs, crimes and lost productivity ~\cite{nida2015}.

These days, people also spend a significant amount of time on social media such as Twitter, Facebook and Instagram to interact with friends and families, exchange ideas and thoughts, provide status updates and organize events and activities. The ubiquity and widespread use of social media underlines the needs to explore its intersection with substance use and its potential as a scalable and cost-effective solution for screening and  preventing substance misuse and abuse. 

In this research, we employ the state-of-the-art machine learning and text mining algorithms to build automated substance use prediction systems, which can be used to identify people who are at risk of SUD. Moreover, by analyzing rich human behavior data on social media,  we can also gain insight into patterns of use and risk factors associated with substance use. The main contributions of this work include:
\begin{enumerate}
\item We have explored a comprehensive set of single-view feature learning methods to take advantage of a large amount of unsupervised social media data. Our results have shown significant improvement over baseline systems that only use supervised training data. 
\item We have explored several multi-view learning algorithms to take advantage of heterogeneous user data such as Facebook ``likes" and ``status updates". Our results have demonstrated significant improvement over baselines that only use a single data type. 
\item We have uncovered new insight into the relationship between a person's social media activities and substance use such as the relationship between word usage and SUD.
\end{enumerate}


\section{Related Work}
Substance use disorder (SUD) encompasses a complex pattern of behaviors. Many studies have been conducted to discover factors interacting with SUD. A growing number of studies have confirmed a strong association between personal traits and substance use. For example, ~\cite{campbell2014personality} found that smokers have significantly higher {\it openness to experience} and lower {\it conscientiousness}, a personality trait related to a tendency to show self-discipline, act dutifully, and aim for achievement.~\cite{cook1998personality} examined the links between alcohol consumption and personality and found that alcohol use is correlated positively with sociability and extraversion. ~\cite{terracciano2008five} conducted a study involving 1102 participants and found a link between drug use and low {\it conscientiousness}. ~\cite{carroll2009modeling} revealed risk factors related to addiction such as age, sex, impulsivity, sweet-liking, novelty reactivity, proclivity for exercise, and environmental impoverishment. Additionally, addiction is also linked to environmental and social factors such as neighborhood environment~\cite{crum1996neighborhood}, family environment~\cite{cadoret1986adoption, brent1995risk} and social norms~\cite{botvin2000preventing, oetting1987common}. 

Traditionally, in behavior science research, data are collected from surveys or interviews with a limited number of people. The advent of social media makes a large volume of diverse user data available to researchers, which makes it possible to study substance use based on online user behaviors in a natural setting.  Typical data from social media include demographics (age, gender etc.), status updates (text posts etc.), social networks (follower and following graph etc.) and  likes (thumb up/down etc.). Recently,   social media analytics has increasingly become a powerful tool to help understand the traits and behaviors of millions of social media users such as personal traits~\cite{golbeck2011predicting, volkova2015predicting, youyou2015computer,kilicc2016analyzing}, brand preferences~\cite{yang2015using}, communities and events~\cite{sayyadi2009event},  influenza trend~\cite{aramaki2011twitter} and crime~\cite{li2012tedas}. So far, however, there has been limited work that directly applies large scale social media analytics to automatically predict SUD.   Among the work known to us, ~\cite{zhou2016understanding} identified common drug consumption behaviors with regard to the time of day and week. They also discovered common interests shared by drug users such as celebrities (e.g, Chris Tucker) and comedians (e.g., cheechandchong). In addition, ~\cite{kosinski2013private} automatically predicted SUD based on social media likes.  Since their dataset is very similar to ours,  we will use the Kosinski model as one of our baselines.

\section{Dataset}
The data for the study was collected from 2007 to 2012 as a part of the myPersonality project~\cite{kosinski2015facebook}. myPersonality was a popular Facebook application that offered to its users psychometric tests and feedback on their scores. The data were gathered with an explicit opt-in consent for reuse for research purposes. Our study uses three separate datasets: Facebook status updates (a.k.a. posts), likes and SUD status. 

The status update dataset contains  22 million textual posts authored by 153,000 users. The average posts per user is 143 and the average words per user is 1730. We removed users who only have non-English posts and those who have written less than 500 words. Our final status update dataset includes 106,509 users with 21 million posts.  After filtering out low frequency words (those appear less than 50 times in our corpus),  the vocabulary size of the status update dataset is 73,935.

The likes data are used by users to express positive sentiment toward various targets  such as products, movies, books, expressions, websites and people (they are called {\it Like Entities} or LEs). Previous studies have demonstrated that social media likes speak volumes about who we are. In addition to directly signaling interests and preferences, social media likes are indicative of ethnicity, intelligence and personality ~\cite{kosinski2013private}.  The like dataset includes the likes of 11 million Facebook users. Overall, there are 9.9 million unique LEs and 1.8 billion user-like pairs. The average likes per user is 161 and the average Likes each LE received is 182. We filter out users as well as LEs  who have a small number of likes. The filtering threshold for users is 50 and is 800 for LEs. After the filtering, our like dataset contains 5,138,857 users and 253,980 unique LEs. 

\begin{table*}[ht]
    \small
    \centering
    \caption{Dataset Descriptions}
     \label{tab:dataset}
    \begin{tabular}{@{}llllll@{}}
      \toprule
      Dataset&users&\multicolumn{1}{c}{AvgUserLikes} & \multicolumn{1}{c}{AvgUserPosts} &\multicolumn{1}{c}{Usage}\\
      \midrule
Likes&5,138,857&184&NA&Single View Feature Learning\\
LikesSUD&3,508&267&NA&Single View SUD Prediction\\
Status&106,509&NA&143&Single View Feature Learning\\
StatusSUD&1,231&NA&195&Single View SUD Prediction\\
LikeStatus&54,757&232&220&Multi-View Feature Learning\\
LikeStatusSUD&896&277&219&Multi-View SUD Predication \\
      \bottomrule
    \end{tabular}
  \end{table*}

The SUD dataset contains a total of 13,557 participants~\cite{stillwell2012effects}. Users were asked to answer questions like ``Do you smoke?", with answers ``daily or more", ``less than daily" or ``never". They also completed the Cigarette Dependence Scale (CDS-5)~\cite{etter2003self},  Alcohol Use Questionnaire(AUQ) ~\cite{townshend2005binge} and the Assessment of Substance Misuse Questionnaire (ASMA)~\cite{willner2000further}.  Based on these assessments,  the participants were divided into groups for each SUD type. For example, based on the assessment of tobacco use, a person is catogorized as ``daily or more" (group 3),  ``less than daily" (group 2), or ``never" (group 1).  The validity of the grouping was confirmed by the CDS-5 scores of the groups. Similarly, based on the assessment of alcohol use, participants were categorized as ``weekly or more" (group 3), ``less than once a week" (group 2) or ``never" (group 1). Finally, based on the assessment of drug use, a person is assigned to ``weekly or more" (group 3), ``less than once a week" (group 2), or ``never" (group 1). Among all the SUD participants, 37\% of them are males and 63\% are females. Their average age is 23 years old.

Since the like, status update and SUD datasets are only partially overlapping, their intersections are usually much smaller. Table~\ref{tab:dataset} summarizes the sizes and usage of these datasets. Table~\ref{tab:SUD} shows additional details of the SUD dataset including the distributions of each SUD class.

In summary, among all the datasets we have, the unsupervised like dataset is the largest (5 million+ people). We also have a significant amount of unsupervised status update data (100k+ users). In contrast, the supervised datasets which have the SUD ground truth are pretty small, ranging from 896 for the  intersection of the likes, status updates and SUD (LikeStatusSUD in Table~\ref{tab:dataset}) to 3508, which is the intersection of the likes and SUD (LikesSUD in Table~\ref{tab:dataset}). Thus, the main focuses of this research  include (1) employing unsupervised feature learning to take advantage of a large amount of unsupervised data (2) employing multi-view learning to combine heterogeneous user data for better prediction.  

 
\begin{table*}[t]
   \small
    \centering
    \caption{Statistics of Different SUD Datasets }
    \label{tab:SUD}
    \begin{tabular}{llllllllllll}
    \toprule
\multirow{2}{*}{Dataset} & \multicolumn{3}{c}{Tabacco Use} & &\multicolumn{3}{c}{Alcohol Use} && \multicolumn{3}{c}{Drug Use} \\\cline{2-4}\cline{6-8}\cline{10-12}
  &	3& 2& 1 &&3	& 2	& 1 && 3	& 2	& 1  \\\hline
LikeSUD		&      498	&290&	2603&&	469&	1174&	1716	&&171	&276	&1965 \\
StatusSUD	&	226	&95&	880	&&     179&	416	& 596	&&76	&102	&671\\
LikeStatusSUD	&	147	&69&        660	&&     123&	290	&453	&&262	&53	&75\\
 \bottomrule
\end{tabular}
\end{table*} 

\section{Single-View Post Embedding (SPE)}
The main purpose of this study is to demonstrate the usefulness of employing unsupervised feature learning to learn a dense vector representation of a user's Facebook posts to take advantage of a large amount of unsupervised data. Since  we only use Facebook status updates (a.k.a. posts) in this study, we call the process Single-view  user Post Embedding (SPE). 
\subsection {Feature Learning Methods}
Since each user is associated with a sequence of textual posts, we have explored the following methods to learn a SPE for the user. 

\textbf{\em {Singular Value Decomposition (SVD)}} is a mathematical technique that is frequently used for dimension reduction~\cite{svd2000}. Given any $m*n$ matrix A, the algorithm will find matrices $U$, $V$ and $W$ such that $A=UWV^{T}$.  Here $U$ is an orthonormal $m*n$ matrix, $W$ is a diagonal $n*n$ metrix and  $V$ is an orthonormal $n*n$ matrix.  Dimensionality reduction is done by computing $R=U*W_r$ where  $W_r$  neglects all but the $r$ largest singular values in the diagonal matrix  $W$.  In our study, the $m$ is the number of users, $n$ is the number of unique words in the vocabulary.  $A_{ij}=k$ where $k$ is how many times $word_{j}$ appears in $user_{i}$'s posts.   

\textbf{{\em Latent Dirichlet Allocation (LDA)}} is a generative graphical model that allows sets of documents to be explained by unobserved latent  topics ~\cite{lda2003}.   For each document, LDA outputs a multinomial distribution over a set of latent topics.  For each topic,  LDA also outputs a multinomial distribution over the vocabulary. 

To learn an SPE for each user based on all his/her posts, we have tried several methods (1) UserLDA: it treats all the posts from each user as one big document and trains an LDA model to drive the topic distribution for this document. The per-document topic distribution is then used as the SPE for this user. (2) PostLDA\_Doc: it treats each post as a separate document and trains an LDA model to derive a topic distribution for each post. To derive the SPE for each user, we aggregate all the per-post topic distribution vectors from the same user by averaging them. (3) PostLDA\_Word: instead of using the $average$ of post-based topic distribution vectors, we used a word-based aggregation method suggested by~\cite{schwartz2013personality}:   
 \begin{align*}
p(topic | user) = \sum_{w\in voc} P(topic | w) * p(w | user)
\end{align*}
where $voc$ represents the vocabulary,  $p(w|user)$ is the probability that word $w$ appears in the posts of $user$ and $p(topic | w)$ is the topic distribution of a word $w$, which is available internally in an LDA model.  For UserLDA model, all the hyper parameters were set to default values. For PostLDA, since Facebook posts are usually short and have a small number of topics in each post, we set the hyper parameter $\alpha$ to 0.3, as suggested by~\cite{schwartz2013personality}

\textbf{\em {Document Embedding with Distributed Memory (D-DM)}} Given a document, D-DM simultaneously learns a vector representation for each word and a vector for the entire document~\cite{doc2vec2014}.  During training, the document vector and one or more word vectors are aggregated to predict a target word in the context.  To learn a SPE for each user,  we have explored two methods (1) User-D-DM: it treats all the posts by the same user as one document and trains a document vector to represent the user. (2) Post-D-DM: it treats each post as a document and train a D-DM to learn a vector for each post.  To derive the SPE for  a user, we aggregate all the post vectors from the same person using ``average". 

  
\textbf{{\em Document Embedding with Distributed Bag of Words (D-DBOW)}} D-DBOW learns a global document vector to predict words randomly sampled from the document. Unlike D-DM,  D-DBOW only learns a  vector for the entire document. It does not learn vectors for individual words. Neither does it use a local context window since the words for prediction are randomly sampled from the entire document.   Similar to D-DM, to derive the SPE for a user, we used two methods (1) User-D-DBOW and (2) Post-D-DBOW.   

\begin{table}
   \small
    \centering
    \caption{SPE: Prediction Results}
    \label{tab:spe_result}
    \begin{tabular}{llll}
\toprule
Methods & Tobacco &Alcohol&Drug \\\hline
Unigram & 0.663&0.672&0.644\\
LIWC&0.731&0.689&0.758\\\hline
SVD&0.779&0.724&0.764\\
UserLDA&0.641&0.603&0.599\\
PostLDA\_Word&0.733&0.617&0.628\\
PostLDA\_Doc&0.768&0.687 &0.721\\
Post-D-DM&0.536&0.622&0.520\\
User-D-DM&0.775&0.730&0.767\\
Post-D-DBOW&0.531&0.606&0.526\\
User-D-DBOW&\textbf{0.802}&\textbf{0.768}&\textbf{0.819}\\
 \bottomrule
\end{tabular}
\end{table}

\subsection {SUD Prediction with SPE}
In our experiments, to search for the best model, we systematically varied the output SPE dimension from 50, 100, 300, to 500.  We used the Gensim implementation of SVD, LDA, D-DM and D-DBOW in our experiments. For D-DM, the context window size was set to 5. 

We compared our models with two baselines that use only supervised learning (1) a unigram model which uses unigrams as the predicting features. Since we have a large number of unigrams,  we performed supervised feature selection to lower the total number of input features. Finally since all our SUD variables have three values, we employed SVM in 3-way classifications.  (2) a LIWC model which uses human engineered LIWC features for SUD prediction.  LIWC is a psycholinguistic lexicon~\cite{pennebaker2015development} that has been frequently used in text-based human behavior prediction. Since the number of LIWC features is relatively small, no feature selection was performed. Here, we only used the {\it Status} dataset in Table~\ref{tab:dataset} as the training data for SPE learning and the {\it StatusSUD} dataset for supervised SUD prediction .

We evaluate the performance of our models  using 10-fold cross validation. The evaluation results shown in Table~\ref{tab:spe_result} are based on weighted ROC AUC of the best models.   Among all the feature learning methods for Facebook status updates, User-D-DBOW performed the best.  It significantly outperformed all the baseline systems that only rely on supervised training ($p<0.01$ based on t-tests). It also significantly outperformed all the traditional feature learning methods such as LDA and SVD ($p<0.01$ based on t-tests).  Moreover, in terms of whether to treat all the posts by the same user as one big document or separate documents,  LDA prefers one post one document (models with a ``post" prefix) while all the document vector-based methods prefer one user one document (models with a ``User" prefix). Moreover, to use post-level LDA to derive the SPE for a user, the document-based aggregation method (PostLDA\_Doc) performed better than the word-based method (PostLDA\_Word).    

\section{Single-View Like Embedding (SLE)}
In addition to textual posts, each user account is also associated with likes.  Since the like dataset is very sparse (e.g., among the millions of unique likes on Facebook, each user only has a small number of likes),  we conduct experiments to learn a dense vector representation for all the likes by a user.  We call this process Single-view user Like Embedding (SLE). 

\subsection{Feature Learning Methods}
The input to SLE is simply a set of LEs liked by a user. Each LE is represented by its id.  To map such a representation to a dense user like vector, we have tried multiple methods.  

\textbf{\em {Singular Value Decomposition (SVD)}} is similar to the one used in SPE except  $A_{ij}=1$ if $user_{i}$ likes $LE_{j}$. Otherwise, it is 0.  Here $A$ is a $m*n$ matrix where  $m$ is the number of users and $n$ is the number of unique LEs in the like dataset. 

\textbf{{\em Latent Dirichlet Allocation (LDA)}}. To apply LDA to the like data, each individual LE is treated as a word token and all the LEs liked by the same person form a document. The order of the LEs in the document is random. For each user, LDA outputs a multinomial distribution over a set of latent ``Like Topics".  For example, a ``Like Topic" about ``hip hop music" may include famous hip hop songs and musicians. 

\textbf{{\em Autoencoder (AE)}} is a neural network-based method for self-taught learning~\cite{autoencoder2006}. It learns an identity function so that the output is as close
to the input as possible. Although an identity function seems a trivial function to learn, by placing additional constraints (e.g,, to make the number of neurons in the hidden layer much
smaller than that of the input), we can still uncover structures in the data. Architecturally, the AE we used has one input layer, one hidden layer and one output layer. For each user,
we construct a training instance $(X, Y)$ where the input vector $X$ and output vector $Y$ are the same. The size of $X$ and $Y$ is the total number of unique LEs in our dataset. $X_{i}$ and $Y_{i}$
equal to 1 if the user likes $LE_i$. Otherwise they are 0.

\textbf{\em {Document Vector with Distributed Memory (D-DM)}} We also applied D-DM to the like data.  Given all the likes of a user, D-DM learns a vector representation for each LE as well as a  document vector for all the LEs from the same user. We use the learned document vector as the output SLE.  
  
\textbf{{\em Document Vector with Distributed Bag of Words (D-DBOW)}} Similarly, we applied D-DBOW to the like dataset. Since D-DBOW  does not use a local context window and the words for prediction are randomly sampled from the entire document, it is more appropriate for the like dataset than D-DM. where the positions of LEs do matter.   


\subsection{SUD Prediction with SLE}
Similarly, we systematically varied the output SLE dimension from 50, 100, 300, to 500 in order to search for the best model.  We used Keras with Theano backend to implement AE. For D-DM, the context window size was set to 20.

We used SVM to perform 3-way classification. We compared our results with a unigram baseline. We also compared our results with the Kosinski model~\cite{kosinski2013private} that was trained on the same like dataset. However, its results were based on two-way classification, a simpler task than 3-way classification.  All the results are based on weighted ROC AUC. 
\begin{table}
   \small
    \centering
    \caption{SLE: Prediction Results}
    \label{tab:sle_result}
    \begin{tabular}{llll}
\toprule
Method & Tobacco &Alcohol&	Drug \\\hline
Unigram&0.687&	0.651&	0.673\\
Kosinski\textsuperscript{$\star$} &$0.730^{*}$ &$0.700^{*}$&$0.650^{*}$ \\\hline
AE &0.678&	0.648&	0.672\\
SVD&0.757&0.756&0.753 \\
LDA&0.723&0.737&0.704 \\
D-DM&0.688&0.713&0.687 \\
D-DBOW&\textbf{0.787}&\textbf{0.795}&\textbf{0.791} \\
 \bottomrule
\multicolumn{4}{l}{$\star$:2-way classification, 3-way for the others} \\
\end{tabular}
\end{table}

As shown in Table~\ref{tab:sle_result}, among all the SLE methods, the D-DBOW model performed the best. It significantly outperformed the unigram baseline that does not use any unsupervised data ($p<0.01$ based on t-tests). It also significantly outperformed all the traditional feature learning method such as SVD and LDA (The Kosinski model used SVD for feature learning) ($p<0.01$ based on t-tests). Between the two document vector-based methods D-DM and D-DBOW, D-DBOW outperformed D-DM. We think this is due to the fact that D-DBOW does not use local context window, thus is not sensitive to the positions of LEs in a document. Since LE positions are randomly decided, D-DBOW seems to be a better fit for the like data.  
\begin{figure*}[t]
\centering
  \centering
  \includegraphics[width=1.0\linewidth]{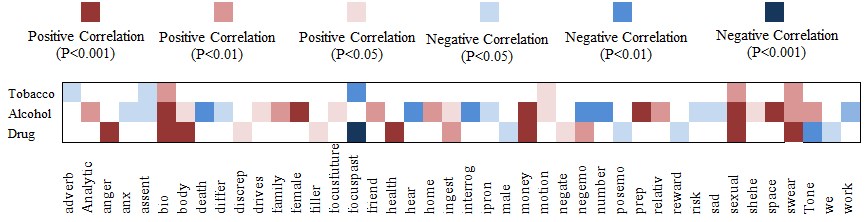}
  \caption{LIWC Features that are Most significantly Correlated Substance Use.}
\label{fig:liwc}
\end{figure*}

\section{Multi-View User Embedding (MUE)}
The main purpose of this study is to demonstrate the usefulness of combining heterogeneous  user data such as likes and posts to learn a dense vector representation for each user. Since we employ unsupervised multi-view feature learning to combine these data, we call this process Multi-view User Embedding (MUE).

\subsection {Feature Learning Methods}
We have explored two multi-view learning algorithms: CCA and DCCA. 

\textbf{\em {Canonical Correlation Analysis (CCA)}} CCA is a statistical method for exploring the relationships between two multivariate sets of variables (vectors)~\cite{hardoon2004canonical}. Given two vectors $X_{1}$ and $X_{2})$, CCA tries to find $w_{1}X_{1}$, $w_{2}X_{2}$ that are maximally correlated: 
\begin{align}
	(w_{1}^{*},w_{2}^{*}) = \operatorname*{arg\,max}_{w_{1},w_{2}}  corr(w_1^{'}X_1,w_2^{'}X_2) \\
	= \operatorname*{arg\,max}_{w_{1},w_{2}}  \frac{w_{1}^{'} \sum_{12}w_{2}}{\sqrt{w_{1}^{'}\sum_{11}w_{1}w_{2}^{'}\sum_{22}w_{2}}}
\end{align}
where $(X_{1}, X_{2})$ denote random vectors with covariances $(\sum_{11},\sum_{22})$ and cross-covariance $\sum_{12}$. CCA has been used frequently in unsupervised data analysis ~\cite{sargin2006multimodal,chaudhuri2009multi,kumar2011co,sharma2012generalized}. 

\textbf{\em {Deep Canonical Correlation Analysis (DCCA)}}
DCCA aims to lean highly correlated deep architectures, which can be a non-linear extension of CCA~\cite{andrew2013deep}. The intuition is to find a maximally correlated representation of the two views by passing them through multiple stacked layers of nonlinear transformation~\cite{andrew2013deep}. Typically, there are three steps to train DCCA: (1) using a denoising autoencoder to pre-train each single view. In our experiments, we pre-train each single view using SPE or SLE. (2) computing the gradient of  the correlation of top-level representation. (3) tuning parameters using back propagation to optimize the total correlation. 

\subsection{SUD Prediction With MUE}
The input to  MUE are the two single views obtained earlier (i.e. SPE or SLE). Here, we choose the outputs from D-DBOW since it consistently outperformed all the other methods in learning SPEs and SLEs.  We have run CCA and DCCA in two settings (1) balanced setting in which the SPE and SLE dimensions are always the same (2) imbalanced setting in which the dimension of SPE may be different from that of SLE.  Since we varied the output dimensions of SPE and SLE  from 50, 100, 300, to 500 systematically, the input dimension to MUE under the balanced setting are 100, 200, 600 and 1000. When running CCA and DCCA under the imbalanced setting, we only chose the best SPE (with 50 dimensions) and the best SLE (with 300 dimensions). We also varied the number of MUE output dimensions systematically from 20,50,100,200,300,400,500 to 1000 (up to the total input MUE dimensions). We used the {\it LikeStatus} dataset in Table~\ref{tab:dataset} as the training data for multi-view unsupervised feature learning. For MUE-based supervised SUD predication, we used the {\it LikeStatusSUD data}. In our experiments, we use the a variant of CCA called {\it wGCCA} implemented by~\cite{benton2016learning} where we set the weights for both views equal~\footnote{https://github.com/abenton/wgcca}. We used the DCCA implementation by ~\cite{andrew2013deep} which uses Keras and Theano  as the deep learning platform~\footnote{https://github.com/VahidooX/DeepCCA}. We also varied the number of hidden layers from 1 to 3 to tune the performance. 

We compared our multi-view learning results with 3 baselines: BestSPE and BestSLE are the best single view models. We also used a 3rd  baseline called {\it Unigram\_combine}, which simply concatenates all the post and like unigrams together and then applies supervised feature selection before uses the remaining features in a SVM-based classification. As shown in Table ~\ref{tab:mue_result},  both wGCCA and DCCA significantly outperformed the unigram-based baseline ($p<0.01$ based on t-test). The difference between the best multi-view models (wGCCA\_balanced for Alcohol and drug, wGCCA\_imbalanced for drugs) and the best single view models are also significant ($p<0.01$).  wGCCA also performed significantly better than DCCA on our tasks ($p<0.01$ based on t-tests).  
\begin{table}
   \small
    \centering
    \caption{MUE: Prediction Results}
    \label{tab:mue_result}
    \begin{tabular}{llll}\hline
    \toprule
	&Tobacco&Alcohol&Drug\\\hline
	BestSPE & 0.802&0.768&0.819\\
	BestSLE & 0.787 & 0.795 & 0.791 \\\hline
	Unigram\_combine&0.685 & 0.669 &0.662 \\\hline
	wGCCA\_balanced&0.848&\textbf{0.811}&\textbf{0.844}\\
	wGCCA\_imbalanced&\textbf{0.855}&0.799&0.832\\
	DCCA\_balanced&0.774&0.778&0.742\\
	DCCA\_imbalanced&0.760&0.781&0.737\\
	\bottomrule
\end{tabular}
\end{table}

\section{Social Media and Substance Use}
In addition to building models that predict SUD, we are also interested in understanding the relationship between a person's social media activities and substance use behavior. Since many of the SPES and SLEs are not easily interpretable, in this section, we focus on the LIWC features from status updates and the LDA topics from both Likes and status updates. Since the SUD ground truth is an ordinal variable and the LIWC/LDA features are numerical, we used Spearman's rank correlation analysis to identify features that are most significantly correlated with SUD. Figure~\ref{fig:liwc} shows the LIWC features that are significantly correlated with at least one type of SUD ($p<0.05$).  The color red represents a positive correlation while blue represents a negative correlation.  In addition, the saturation of the color indicates the significance of the correlation. The darker the color is, the more significant the correlation is. 
\begin{table*}[t]
\small
\centering
    \caption{Topics Most Significantly Correlated with Substance Use.}
    \label{tab:lda}
    \begin{tabular}{@{}lll@{}}
      \toprule
       & Significance & Topic \\
      \midrule
       \textbf{Tobacco}\\
       \midrule
	  \multirow{2}{*}{Posts} & + &	{\it (T1) fuck, shit, ass, fucking, bitch, face, don't, kick, damn, man, lol, hell...}\\
	      & - &{\it (T2) paper, book, writing, read, class, essay, english, finished, reading, time, page ...}\\\hline
	 \multirow{2}{*}{Likes} &+& {\it (T3) Tool, Misfits, A Perfect Circle, Rob Zombie ...	
} 

\\
	 &-& {\it (T4) The Twilight Saga, Forever 21, Twilight, Victoria's Secret, Katy Perry}\\
       \midrule
        \textbf{Alcohol}\\
       \midrule
       \multirow{2}{*}{Posts}&+&{\it (T5)  tonight, night, free, party, tickets, bar, saturday, friday, dj, drink, club, show, beer, ladies...
}\\
       &--&{\it (T6) class, history, paper, math, science, writing, essay, finished, study, test, final, exam ...
}\\\hline
	  \multirow{2}{*}{Likes}&++&{\it  (T7) V For Vendetta, Boondock Saints,  Pan's Labyrinth ...}\\      
	  &--& {\it (T8) Cookie Monster, Squirt, Last Day of School, Hunger Games Official Page, Wonka ...}\\
	  \midrule
        \textbf{Drug}\\
       \midrule
       	\multirow{2}{*}{Posts}& ++ & {\it (T9)  fuck, shit, ass, fucking, bitch, face, don't, kick, damn, man, lol, hell...}\\
        & - &{\it (T10) dinner, nice, shopping, christmas, home, weekend, lunch, family, house,love,wine :-)...
}\\\hline
      \multirow{2}{*}{Likes}&+&{\it (T11) Radiohead,The Cure, Depeche Mode, The Smiths, Arctic Monkeys ...}\\
	  & - & {\it (T12) Music, Movies, Traveling, Photography, Dancing ...}\\
      \bottomrule
    \end{tabular}
\end{table*} 

As shown in Figure~\ref{fig:liwc}, swear words such as ``fuck" and ``shit", sexual words such as ``horny" and ``sex", words related to biological process such as ``blood" and ``pain" are positively correlated with all three types of SUD.  In addition, words related to money such as ``cash" and ``money", words related to body such as ``hands" and ``legs", words related to ingestion such as ``eat" and ``drink" are positively correlated with both alcohol and drug use; words related to motion such as ``car" and ``go" are positively correlated with both alcohol and tobacco use. In addition, female references such as ``girl" and ``woman", prepositions, space reference words such as ``up" and ``down" are positively correlated with alcohol use, while words related to anger such as ``hate" and ``kill", words related to health such as ``clinic" and ``pill" are positively correlated with drug use. 

In terms of LIWC features that are negatively correlated with SUD, words associated with the past such as ``did" and ``ago" are negatively correlated to both tobacco and drug use; assent words such as ``ok", ``yes" and ``agree" are negatively correlated to both alcohol and tobacco use.  In addition, male references such as ``boy" and ``man", words related to reward such as ``prize" and ``benefit", words related to positive emotions such as ``nice" and ``sweet", first person pronouns (plural) such as ``we" and ``our" are negatively correlated to drug use. Moreover, impersonal pronouns such as ``it", differentiation words such as ``but" and ``else", and work-related words such as ``job" and ``work" are negatively correlated with alcohol use. Surprisingly, risk related words such as ``danger", words related to sadness, death and negative emotions are also negatively correlated with alcohol use. 

There are a few surprising correlations in our results. For example, female references such as ``girl" and ``woman" are positively related to alcohol use while male references such as ``man" and ``boy" are negatively related to drug use.  To interpret this, previous research has shown~\cite{schwartz2013personality} that female references actually are used more often by male authors and vice versa. Thus, our findings suggest that males are more likely to use alcohol while females are less likely to use drugs.  

We have also used Spearman's correlation analysis to identify SUD-related ``Like Topics" and ``Status update Topics" learned by LDA. Since the number of significant topics is quite large, in Table~\ref{tab:lda}, we only show a few samples.  From the table, we can see based on a user's status updates,  ``swear topics" (T1, T9) are positively correlated with both tobacco and drug use, which is consistent with our LIWC findings.  The ``night life topic" (T5) is positively related to alcohol use. In addition, school related topics (T2, T6) are negatively correlated with tobacco and alcohol use. Positive family-related activities (T10) are negatively correlated with drug use.  In addition, based on the LDA topics learned from ``like",  a preference for rock music (T3,T11) is positive correlated with tobacco and drug use. A preference for movies such as ``V For Vendetta" and ``Boondock Saints" (T7) is positively correlated with alcohol use, while having a hobby (T12),  liking cartoons and shows favored by kids (T8) or  liking movies and brands favored by girls (T4) are negatively correlated with drug, alcohol and tobacco use respectively. 

\section{Discussion and Future Work}
Currently, our multi-view unsupervised features learning methods only learn from the intersection of the like and status update data, which is much smaller than either the like or the status update data.  Similarly, MUE-based supervised prediction used only  the intersection of all three datasets which is very small (only contains 896 users). Thus,  it would be useful if a future multi-view feature learning algorithm is capable of using all the available data (e.g., the union of all the supervised and unsupervised training data).  Moreover, our best SPE model only has 50 dimensions while our best SLE model has 300 dimensions. This  might be because the supervised training data used by SPE is almost three times smaller than that for SLE . But surprisingly, SPE-based models performed better than SLE-based models. We expect that with more training data, the performance of SPE-based methods can be further improved.
 
\section{Conclusion}
We believe social media is a promising platform for both studying SUD-related human behaviors as well as engaging the public for substance abuse prevention and screening.  In this study, we have focused on four main tasks (1) employing unsupervised features learning to take advantage of a large amount of unsupervised social media user data (2) employing multi-view  feature learning to combine heterogeneous user information such as "likes" and "status updates" to learn a comprehensive user representation (3) building SUD prediction models based on learned user features (4) employing correlation analysis to obtain human-interpretable results. Our investigation has not only produced models with the state-of-the-art prediction performance (e.g., for all three types of SUD, our models achieved over 80\% prediction accuracy based on AUC) , but also demonstrated the benefits of incorporating unsupervised heterogeneous user data for SUD prediction. 


\bibliography{emnlp2017}
\bibliographystyle{emnlp_natbib}

\end{document}